\documentclass[11pt]{article}

\usepackage{acl}
\usepackage{booktabs}   

\usepackage{times}
\usepackage{latexsym}
\newcommand\blfootnote[1]{%
  \begingroup
  \renewcommand\thefootnote{}\footnote{#1}%
  \addtocounter{footnote}{-1}%
  \endgroup
}

\usepackage[T1]{fontenc}

\usepackage[utf8]{inputenc}

\usepackage{microtype}

\usepackage{inconsolata}

\usepackage{graphicx}
\usepackage{array}
\usepackage{algorithm}
\usepackage{amsmath}
\usepackage{tabularx}
\usepackage{booktabs}
\usepackage{url}

\usepackage{algpseudocode}
\newcommand{\StepHeader}[1]{\Statex \textbf{#1}}
\usepackage{listings}
\lstdefinestyle{prompt}{
basicstyle=\footnotesize\ttfamily,
breaklines=true,
breakatwhitespace=false,
columns=flexible,
keepspaces=true,
backgroundcolor=\color{gray!7},
frame=single,
framesep=6pt,
rulecolor=\color{gray!40},
xleftmargin=6pt,
xrightmargin=6pt,
aboveskip=4pt,
belowskip=4pt,
}

\title{\textsc{CrowdMath}: A Dataset of Crowdsourced\\
Mathematical Research Discussions}

\author{First Author \\
  Affiliation / Address line 1 \\
  Affiliation / Address line 2 \\
  Affiliation / Address line 3 \\
  \texttt{email@domain} \\\And
  Second Author \\
  Affiliation / Address line 1 \\
  Affiliation / Address line 2 \\
  Affiliation / Address line 3 \\
  \texttt{email@domain} \\}

\author{
 Sherin Muckatira\textsuperscript{1,\textdagger,*} \quad
  Jesse Geneson\textsuperscript{2, \textdagger} \quad
  Slava Gerovitch\textsuperscript{3} \quad
  Pavel Etingof\textsuperscript{3} \\
  \textbf{Mikhail Gronas\textsuperscript{4}} \quad
  \textbf{Anna Rumshisky\textsuperscript{1,5, \ensuremath{\diamond}}} \\
  \textsuperscript{1}University of Massachusetts Lowell \quad
  \textsuperscript{2}San Jose State University \quad
  \textsuperscript{3}Massachusetts Institute of Technology \\
  \textsuperscript{4} Dartmouth College\quad
  \textsuperscript{5}Amazon AGI
}
\begin{document}
\maketitle
\blfootnote{\raggedright\textsuperscript{\textdagger}\,Equal contribution.}
\blfootnote{\raggedright\textsuperscript{*}\,Corresponding author:\ \texttt{sherinbojappa\_muckatira@student.uml.edu}}
\blfootnote{\raggedright\textsuperscript{\ensuremath{\diamond}}\,Work done at University of Massachusetts Lowell.}
\begin{abstract}
Large language models have made substantial progress on mathematical reasoning, but existing benchmarks typically evaluate well-specified problems with final answers, step-by-step solutions, or complete proofs. They do not capture collaborative open-problem solving: a setting in which participants propose partial arguments, identify gaps or errors in prior steps, repair flawed reasoning, and gradually synthesize incremental contributions into a proof.  We introduce \textsc{CrowdMath}, a dataset of 164 expert-annotated progress chains from the MIT PRIMES--Art of Problem Solving (AoPS) CrowdMath program (2016--2025), a collaborative research initiative whose discussions have led to peer-reviewed publications. Each chain traces a multi-participant forum discussion from an open-problem statement to a completed proof. Posts are labeled by their functional roles in the evolving solution process, including partial progress, proof completion, erroneous reasoning, and error identification. We define evaluation tasks and benchmark six frontier models. Models achieve 83--88\% accuracy on next-post prediction, suggesting that they can follow the local flow of mathematical discussion. However, they struggle to identify the functional significance of individual contributions with the best model achieving only 0.42 macro-F1 on post-role classification. \textsc{CrowdMath} exposes a gap between solving well-specified mathematical problems and understanding collaborative mathematical progress as it unfolds.
\end{abstract}

\section{Introduction}

\begin{figure*}[!t]
\centering
\includegraphics[
  width=0.8\textwidth,
  trim=40 0 0 0,
  clip
]{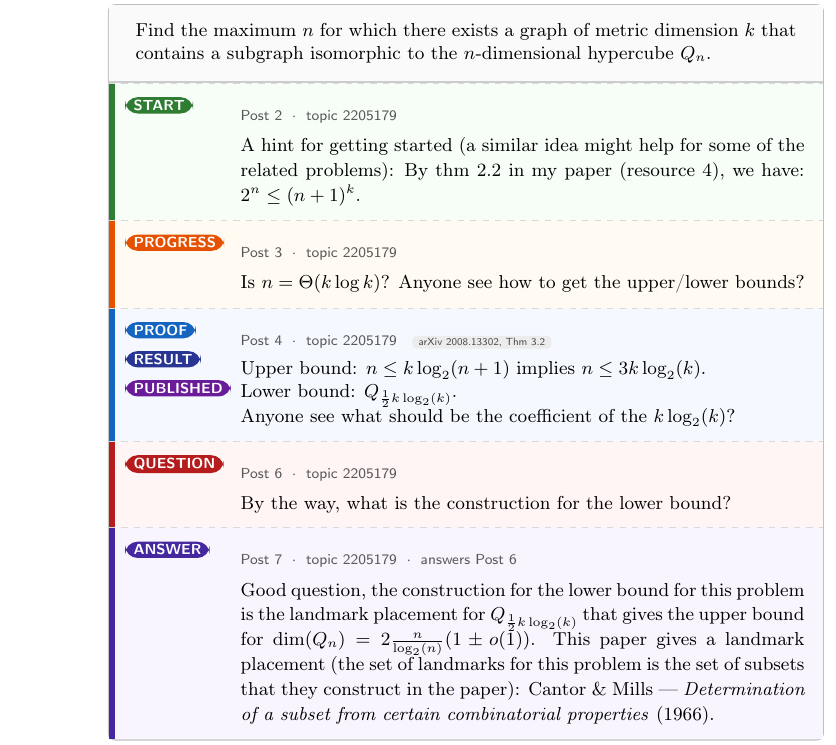}

\caption{Example progress chain from \textsc{CrowdMath}. The chain shows how an open problem discussion develops, including the start, partial progress, proof completion, and a follow-up question--answer exchange. The example is drawn from the metric-dimension and hypercube subgraph thread for Open Problem 2020-6.
}
\label{fig:chain-example}
\end{figure*}

Large language models (LLMs) have demonstrated impressive mathematical capabilities, with strong performance on grade-school word problems, competition-style problem solving, and formal theorem proving~\citep{cobbe2021trainingverifierssolvemath,hendrycksmath2021,yang2023leandojo}. These datasets and benchmarks have been valuable for measuring whether models can solve well-specified problems by producing a correct final answer, derivation, or machine-verifiable proof. Even when prior work exposes intermediate reasoning steps, it typically studies reasoning within a single solution attempt. This leaves open a complementary setting: collaborative open-problem solving, where mathematical progress is distributed across multiple posts and participants.

In collaborative open-problem solving, individual contributions may be incomplete in isolation but still essential to the eventual solution. A participant may propose a partial idea, test an example, identify an error, refine a conjecture, ask or answer a clarifying question, or connect earlier observations into a proof path. Understanding such discussions therefore requires more than checking whether a final answer is correct. A model must track where the discussion currently stands, how one contribution relates to earlier ones, whether a proposed step is valid or useful, and what kind of contribution would advance the argument.

We introduce \textsc{CrowdMath}, a dataset of expert-annotated collaborative mathematical research discussions drawn from the MIT PRIMES--Art of Problem Solving (AoPS) CrowdMath program \footnote{\url{https://artofproblemsolving.com/polymath}}. \textsc{CrowdMath} contains 164 progress chains tracing the development from open-problem statements to completed proofs across multi-participant forum threads. Each post is annotated with its functional role in the evolving solution process, including \textsc{Progress}, \textsc{Proof}, \textsc{Erroneous}, \textsc{FindError}, \textsc{Question}, and \textsc{Answer}, with links to relevant prior posts and results.

\textsc{CrowdMath} differs from prior resources in three key ways. First, it captures real-world research trajectories rather than textbook exercises or synthetic problems. Second, it annotates the process of mathematical progress, including partial ideas, errors, corrections, questions, answers, and proof-completing steps. 
Third, it preserves the collaborative structure of these open-problem discussions, enabling evaluation of whether models can track dependencies across participants and reason about how individual posts contribute to a shared mathematical result.
\blfootnote{\raggedright Code and the dataset are released at \url{https://github.com/text-machine-lab/crowdmath}}

We evaluate frontier models on tasks derived from these annotations, including post-role classification and next-post prediction. Our results show that models can often follow the surface flow of mathematical discourse, but struggle to identify the functional significance of individual contributions, especially when distinguishing proof-completing posts from partial progress. These findings suggest that current models remain limited in their ability to understand mathematical progress as an evolving collaborative process.

Our main contributions are as follows:
\begin{enumerate}
\item A dataset \textsc{CrowdMath} of naturally occurring, multi-participant open mathematical problem-solving discussions from the MIT PRIMES CrowdMath program, with post-level annotations capturing each
contribution's role in the evolving solution process.
\item Evaluation tasks that test whether frontier models can classify proof steps and predict next contributions.
\item Baseline results showing that frontier models struggle to
distinguish proof completions from partial progress, exposing a gap
in process-level mathematical understanding.
\end{enumerate}

\section{Related Work}

\begin{table*}[t]
\centering
\begin{tabular}{>{\ttfamily}lp{9.5cm}}
\toprule
\normalfont Label & Definition \\
\midrule
Proof(y)          & $S_i$ completes the first accepted proof of result $y$.\\[2pt]
NewProof(y)     & $S_i$ completes a better proof of $y$ than the most recent
                    accepted proof.\\[2pt] 
Start(y)          & $S_i$ initiates the thread of discussion leading to the
                    first proof of $y$. \\[2pt]
Progress(y)     & $S_i$ contributes to the proof of $y$ (may be an idea, lemma, or
                    question). \\[2pt]
NewProgress(y)     & $S_i$ contributes to new progress of result $y$ than the most recent accepted proof.\\[2pt]                    
Question(y)       & $S_i$ asks a useful question relevant to result $y$. \\[2pt]
Answer(y,j)       & $S_i$ answers question in $S_j$ relevant to result $y$. \\[2pt]
Erroneous(y)      & $S_i$ claims a proof of result $y$ with a significant
                    error. \\[2pt]
FindError(y,j)    & $S_i$ finds an error in a claimed proof of result $y$ in
                    $S_j$. \\[2pt]
Result                  & A new mathematical finding not already known in the
                          literature. Clarifications, restatements of known
                          theorems, or confirmations of known results are not
                          considered new results. \\[2pt]
Published-in-paper      & Whether the results have been published in a paper;
                          carries the arXiv ID as argument. \\[2pt]
Theorem-number          & Theorem number assigned in the published paper. \\[2pt]
prev-ref    & Post identifier of the previous post contributing to
                          this result. \\
                    
\bottomrule
\end{tabular}
\caption{\textsc{CrowdMath} annotation schema for collaborative open-problem solving. Post-level labels capture functional roles such as proposing problems, making progress, completing proofs, errors, and identifying errors. Metadata fields record publication links, theorem numbers, and dependencies on prior posts.}
\label{app:schema}
\end{table*}

\paragraph{Grade School and Competition Style benchmarks}
There have been multiple math datasets that evaluate mathematical reasoning in language models with closed-form answers. Datasets such as \citep{ling-etal-2017-program, amini-etal-2019-mathqa, cobbe2021trainingverifierssolvemath, miao-etal-2020-diverse} focus on grade-school and university-level mathematics with most of them having natural language rationale and a closed-form answer. 
Datasets such as \citep{hendrycksmath2021, gao2025omnimath, he-etal-2024-olympiadbench, sun2026omega, yue2024harpchallenginghumanannotatedmath, mao-etal-2024-champ, gulati2025putnamaxiom} have harder settings that include high-school level competitions,  Olympiad-style problems, and university-level contest problems. Dynamic and synthetic competition-style benchmarks further expand coverage by adding newly sourced or generated problems \citep{balunovic2026matharena, manem2025sandmathusingllmsgenerate}.  While these datasets and benchmarks  are well suited to problems with known solutions, they are less representative of complex open problems in mathematics. CrowdMath tries to capture the steps involved in solving open problems.

\paragraph{Research Level Math Reasoning datasets} 
Recent datasets move beyond school and competition mathematics toward research-level reasoning.  \citet{glazer2025frontiermathbenchmarkevaluatingadvanced} introduce expert-crafted problems designed to be difficult for frontier models, while \citet{zhang2026realmath} draw problems from research papers and mathematical forums. Other works focus on open problems in specific mathematical areas, such as algebraic combinatorics and computational or applied mathematics \citep{chau2025machine, wang2026horizonmathmeasuringaiprogress}. These benchmarks capture whether a model can arrive at a correct final answer, but not the process involving mathematical progress. CrowdMath captures how mathematicians collectively make progress on open problems.

\paragraph{Process supervision and proof-oriented datasets.}
Another set of datasets studies mathematical reasoning through process supervision, where supervision is provided at the level of intermediate reasoning steps rather than only the final answer. These datasets  \citep{lightman2024lets, wang-etal-2024-math} annotate individual steps in a solution with labels indicating  correctness, using either human annotation or automatically generated supervision. Process-supervision datasets typically focus on evaluating the correctness of steps within a single solution trajectory. In contrast, our dataset captures open mathematical problem solving as a broader collaborative process, where progress may arise from partial ideas, failed attempts, and corrections across multiple contributors. Formal and natural-language theorem-proving benchmarks evaluate whether models can prove mathematical statements, fill in missing proof steps, or translate informal mathematics into formal systems \citep{wellecknaturalproofs, zheng2022miniff, liu2023fimochallengeformaldataset, azerbayev2023proofnet, li2021isarstep, yang2023leandojo}.  While formal theorem-proving datasets provide rigorous benchmarks for machine-checkable reasoning, they do not capture the informal, exploratory, and collaborative processes that often characterize progress on open mathematical problems.

  \begin{figure*}
    \centering
    \includegraphics[width=0.75\textwidth]
    {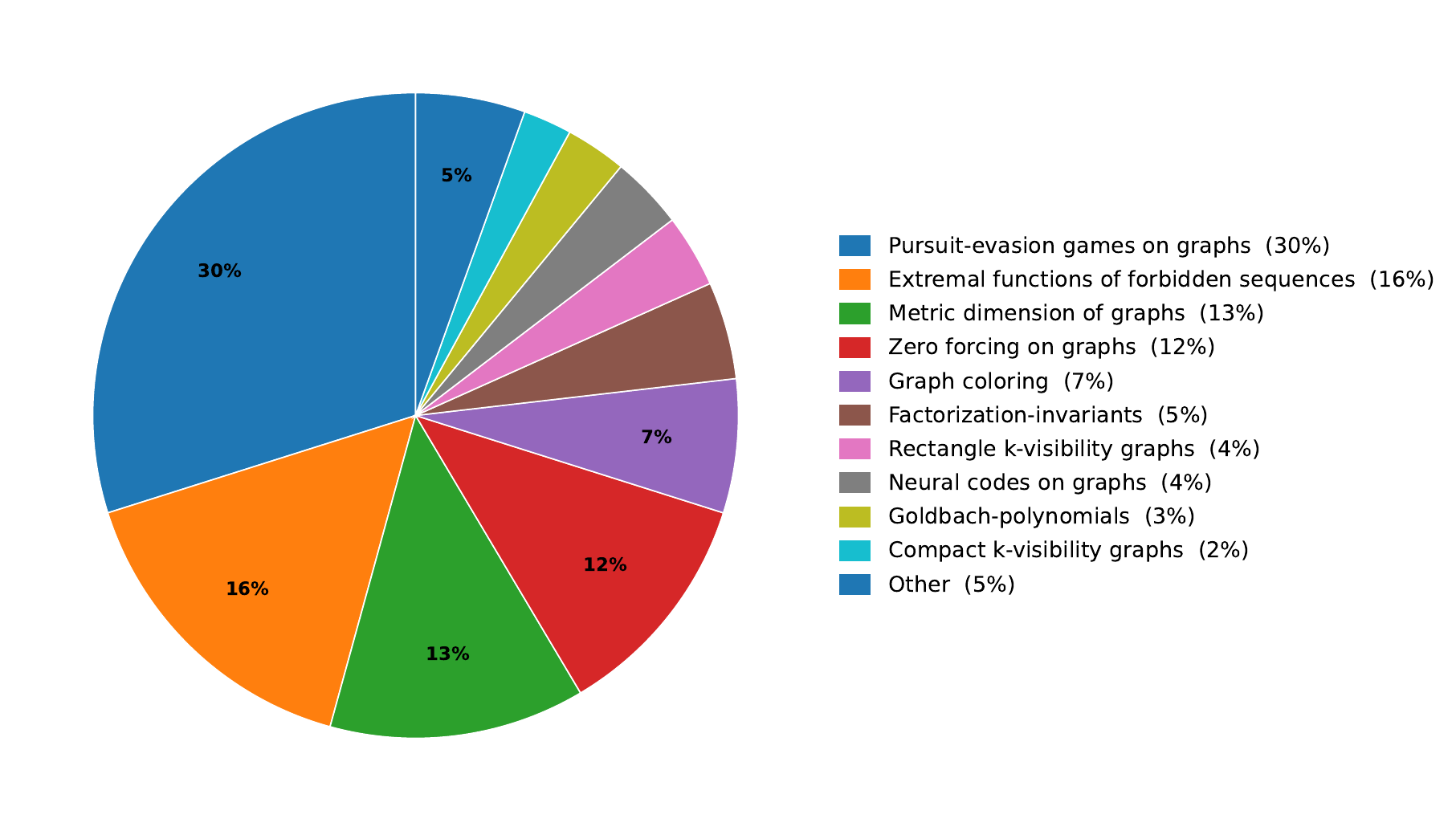}
    \caption
      {Distribution of the 164 \textsc{CrowdMath} progress chains across mathematical topics. The largest areas are pursuit--evasion games on graphs, extremal functions of forbidden sequences, metric dimension of graphs, and zero forcing on graphs. Topics contributing fewer than 2\% of chains are grouped as ``Other.''}
    \label{fig:posts}
  \end{figure*}

\paragraph{Collaborative mathematical discourse.}
A smaller body of work studies mathematics as dialogue. \citet{macina-etal-2023-mathdial} introduce teacher-student dialogue dataset between a human teacher and an LLM student with a focus on common student errors. \citet{suresh-etal-2022-talkmoves} study classroom mathematics discussions through annotated K-12 lesson transcripts from video recordings. Unlike these datasets which study mathematical dialogue in educational settings, CrowdMath targets collaborative open problem solving and captures informal, multi-contributor discussions around difficult mathematical problems, where progress is often partial, non-linear, and distributed across participants.

\paragraph{Creativity and open-ended mathematical problem solving.}
Recent work has also begun to evaluate creativity in mathematical reasoning. \citet{ye2025assessing} study whether models can propose innovative solutions after seeing known solutions, and \citet{chen2025deepmathcreativebenchmarkevaluatingmathematical} evaluate models on novel problems across areas such as algebra, geometry, and analysis. These works focus on assessing the creativity of model-generated solutions. CrowdMath provides a dataset for studying the collaborative process of open problem solving, where progress may emerge through partial ideas, errors, corrections, and counterexamples across multiple contributors.

\section{CrowdMath Program} 

CrowdMath is a large-scale collaborative mathematics research program launched in 2016 by MIT PRIMES and the Art of Problem Solving (AoPS), inspired by the Polymath projects of Timothy Gowers and Terence Tao. It is open-enrollment and globally accessible, allowing high school and undergraduate participants to contribute without a formal selection process. Each annual cycle focuses on a small collection of original, unsolved problems, typically in combinatorics, graph theory, and related discrete areas, chosen to be approachable yet research-level. Participants work through them on persistent public AoPS forum threads under the guidance of mentors. Several projects have produced peer-reviewed publications in venues such as the Electronic Journal of Combinatorics and Discrete Applied Mathematics.

From a data perspective, CrowdMath provides an unusually rich record of mathematical discovery. Unlike curated problem sets or competition archives, the forum logs capture the full temporal evolution of reasoning: exploratory computations, partial proofs, counterexamples, errors, and corrections, all embedded in multi-participant discourse and tied to verifiable published outcomes. This combination of authentic open-problem research, collaborative interaction, and explicit intermediate states makes CrowdMath particularly well-suited for constructing datasets that target fine-grained reasoning, error detection, and process-level supervision rather than final-answer accuracy alone.


\begin{table*}[t]
  \centering
  \small
  \setlength{\tabcolsep}{8pt}
  \begin{tabular}[t]{lr}
  \toprule
  \multicolumn{2}{c}{\textit{(A) Chains per year}} \\
  \midrule
  \textbf{Year} & \textbf{Chains} \\
  \midrule
  2016 & 26 \\
  2017 & 59 \\
  2018 &  6 \\
  2019 & 11 \\
  2020 & 21 \\
  2021 & 19 \\
  2022 &  8 \\
  2023 &  2 \\
  2024 &  5 \\
  2025 &  7 \\
  \midrule
  Total & 164 \\
  \bottomrule
  \end{tabular}
  \hspace{2em}
  \begin{tabular}[t]{lrr}
  \toprule
  \multicolumn{3}{c}{\textit{(B) Posts per result}} \\
  \midrule
  \textbf{Posts} & \textbf{Chains} & \textbf{\%} \\
  \midrule
  1        & 81 & 49.4 \\
  2        & 22 & 13.4 \\
  3        & 25 & 15.2 \\
  4        &  9 &  5.5 \\
  5        &  8 &  4.9 \\
  6--10    & 15 &  9.1 \\
  $\geq$11 &  4 &  2.4 \\
  \midrule
  Mean     & \multicolumn{2}{r}{2.7} \\
  Median   & \multicolumn{2}{r}{2}   \\
  Max      & \multicolumn{2}{r}{25}  \\
  \bottomrule
  \end{tabular}
  \hspace{2em}
  \begin{tabular}[t]{lrr}
  \toprule
  \multicolumn{3}{c}{\textit{(C) Label distribution}} \\
  \midrule
  \textbf{Label} & \textbf{Count} & \textbf{\%} \\
  \midrule
  Proof       & 164 & 30.5 \\
  Result       & 164 & 30.5 \\
  Progress    &  28 & 5.2 \\
  Question    &  42 & 7.8 \\
  Start       &  60 & 11.2 \\
  Answer      &  37 & 6.9 \\
  Erroneous   &  16 &  3.0 \\
  FindError   &  16 &  3.0 \\
  NewProgress &   9 &  1.7 \\
  NewProof    &   2 &  0.4 \\
  \midrule
  Total & 538 & \\
  \bottomrule
  \end{tabular}
    \caption{Dataset statistics for \textsc{CrowdMath}. 
(A) Number of annotated progress chains by year from 2016--2025. 
(B) Distribution of chain lengths, measured by the number of posts associated with each completed result. 
(C) Number of chains containing each label type.}
  \label{tab:dataset-stats}
  \end{table*}

\section{Dataset} 

\subsection{Annotation Process}
\label{sec:annotation-process}

Each post $S_i$ is assigned one or more labels from the schema in Table~\ref{app:schema}. Because most labels are result-centric (taking a result identifier $y$, and sometimes a post identifier $j$, as arguments) a post cannot be fully labeled until the results it references have been identified. We therefore annotate each CrowdMath thread in two passes.

\textbf{Pass 1: Result identification.}
We read the thread to identify every result established in its discussion, where a result is a new mathematical finding as defined in Table~\ref{app:schema}. Each result is given a unique identifier derived from its completion post: a result completed in post $k$ of thread $X$ is labeled $X$-$k$, with a trailing letter appended when a single post completes multiple results (e.g., $X$-$k\mathrm{a}$, $X$-$k\mathrm{b}$). When a result corresponds to a theorem in a published CrowdMath paper, we align the two and record the associated metadata (\texttt{Published-in-paper}, \texttt{Theorem-number}).
\newline

\textbf{Pass 2: Post labeling.}
With results indexed, we revisit the thread and assign post-level labels. Labels such as \texttt{Progress}($y$), \texttt{Proof}($y$), and \texttt{Erroneous}($y$) reference a result $y$; labels such as \texttt{Answer}($y,j$) and \texttt{FindError}($y,j$) additionally reference a prior post $S_j$. 
\newline

\textbf{Cross-thread results:}
A post in one thread occasionally contributes to a result completed in a different thread. Since result identifiers are determined by completion posts, labels referencing such a result cannot be finalized until the thread containing its completion has been processed. We maintain a global result registry across threads and defer finalization of these labels until the referenced completion post is indexed, ensuring all references are globally consistent.

\textbf{Converting raw annotations into progress chains:} The raw annotations consist of post-level labels that may reference
results and, in some cases, prior posts. We convert these annotations
into progress chains by first identifying the set of mathematical results
established in each thread, then collecting all posts whose labels refer
to each result. Each resulting chain contains the open-problem statement,
any associated setup or resource text, and the sequence of posts that
contribute to the corresponding result, ordered by post number. This
construction allows the dataset to represent a proof trajectory rather
than a flat collection of labeled posts. The full chain-construction
algorithm is given in Appendix~\ref{app:algo}, and the post-level label
schema is shown in Table~\ref{app:schema}.

\begin{figure}
    \centering
    \includegraphics[width=0.5\textwidth]
    {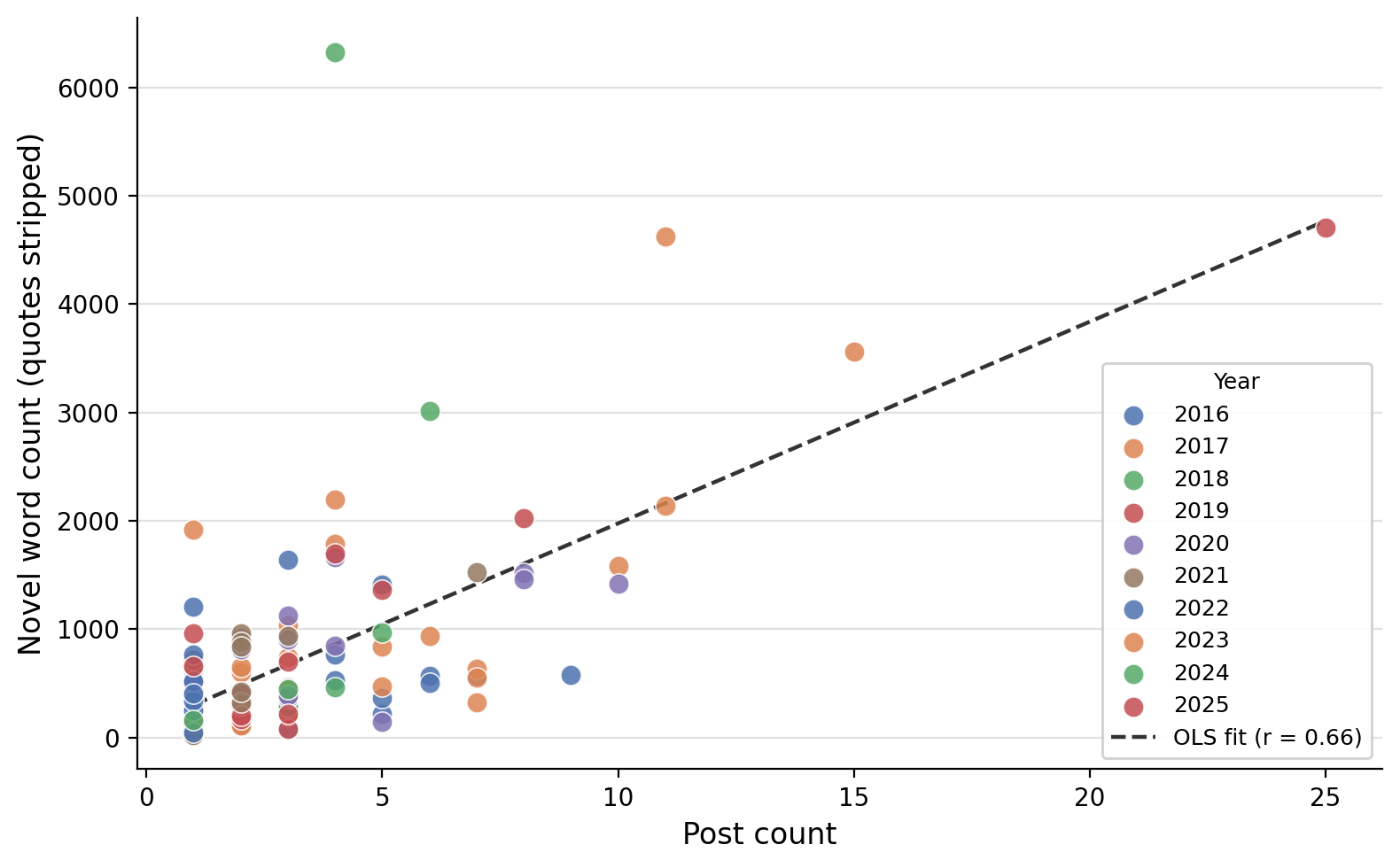}
    \caption{Post count versus text volume for \textsc{CrowdMath} progress chains. Each point represents one completed result. The x-axis shows the number of posts in the chain, and the y-axis shows the total number of novel words after removing quoted prior-post content. Point color indicates the year of the completed result.}
    \label{fig:posts_vs_word}
  \end{figure}

\paragraph{Anonymization.}
Before we received the data, AoPS performed an initial metadata anonymization step: the dataset was provided as a CSV file with one row per post, in which the author username field was replaced by a hash string and exact timestamps were removed, with only chronological post order retained. Because this process anonymized metadata fields but did not systematically anonymize the contents of posts, some post bodies still contained names or other identifying references. We therefore performed an additional systematic scan of the post text to identify and replace remaining personally identifiable information, including real names, forum usernames, and other identifying references.

\subsection{Dataset Statistics}

\textsc{CrowdMath} contains 164 annotated progress chains drawn from
MIT PRIMES CrowdMath discussions between 2016 and 2025. The number of
chains varies substantially by year. Table~\ref{tab:dataset-stats}(A) shows the number of chains per year. The
largest share of chains comes from 2017, with 59 chains, followed by
2016 with 26 chains, 2020 with 21 chains, and 2021 with 19 chains.
Later years contribute fewer chains, with 2023, 2024, and 2025
contributing 2, 5, and 7 chains respectively.

The dataset covers a range of topics in discrete mathematics and
combinatorics. As shown in Figure~\ref{fig:posts}, the
largest topic area is pursuit-evasion games on graphs, accounting for
30\% of progress chains. Other major areas include extremal functions
of forbidden sequences (16\%), metric dimension of graphs (13\%), zero
forcing on graphs (12\%), graph coloring (7\%), and
factorization-invariants (5\%). This distribution reflects the research
focus of the CrowdMath program, where problems are chosen to be
accessible to participants while still leading to publishable
mathematical results.

Progress chains vary in length. Nearly half of the chains contain a
single post associated with the completed result, 81 out of 164 chains
(49.4\%). Table~\ref{tab:dataset-stats}(B) shows a detailed breakdown of the chain lengths. A substantial portion require multiple posts. The mean
chain length is 2.7 posts, the median is 2 posts, and the longest chain
contains 25 posts. This shows that \textsc{CrowdMath} contains both
compact proof trajectories and longer multi-step discussions in which
progress accumulates across several contributions.

Every chain contains a \textsc{Proof} label and a corresponding
\textsc{Result} entry as the dataset is constructed around completed results, giving 164 instances of each. Intermediate
contributions are also common. As shown in Table~\ref{tab:dataset-stats}(C), \textsc{Progress} appears in 28 chains times,
\textsc{Question} 42 times, \textsc{Start} 60 times, and
\textsc{Answer} 37 times. Error-related labels are less frequent but
important for evaluating reasoning over flawed arguments:
\textsc{Erroneous} appears 16 times and \textsc{FindError} appears 16
times, \textsc{NewProgress} appears 9 times and \textsc{NewProof} appears 2 times. Each label type is counted only once in a chain. These labels allow the dataset to represent not only successful
proof completion, but also partial progress, clarification, and error
correction.

Figure~\ref{fig:posts_vs_word}
plots the number of posts in each chain against the total number of
novel words per proved result after removing quoted prior-post content. Each point in the scatter plot represents a result with the color indicating the year in which the result was completed. Most progress chains are compact with fewer posts and fewer novel words. However, there are some progress chains with few posts but a large amount of novel mathematical text, indicating denser proof contributions rather than extended back-and-forth discussion.



\section{Evaluations}
 We evaluate six
frontier models (GPT-5.4, Claude Opus 4.6, Grok 4.20, DeepSeek V3.2, Gemini 2.5 Flash, Qwen3-235B) at temperature 0 and other default settings in OpenRouter\footnote{\url{https://openrouter.ai/docs/quickstart}} for these models. The model details can be found in the appendix table   \ref{tab:model_details}. The prompts used for this task are in Appendix \ref{app:prompts}. 
\label{sec:benchmark}

\begin{table*}[t]
\centering

\begin{minipage}[t]{0.60\textwidth}
\centering
\footnotesize
\textbf{(a) Task 1: Post-role classification}

\vspace{3pt}

\begin{tabular}{@{}lcc@{}}
\toprule
\textbf{Method / Model}
& \textbf{Macro-F1 $\uparrow$}
& \textbf{Accuracy $\uparrow$} \\
\midrule

\multicolumn{3}{@{}l}{\textit{Baselines}} \\
Majority class (\textsc{Result})
    & \textbf{0.187}
    & \textbf{0.596} \\
Position-only 
    & 0.186
    & 0.589 \\

\midrule
\multicolumn{3}{@{}l}{\textit{Frontier models}} \\
GPT-5.4
    & 0.362 [0.297, 0.418]
    & 0.316 [0.262, 0.374] \\
Claude Opus 4.6
    & \textbf{0.424} [0.357, 0.483]
    & 0.371 [0.313, 0.429] \\
Grok 4.20
    & 0.396 [0.326, 0.459]
    & \textbf{0.389} [0.331, 0.447] \\
DeepSeek V3.2
    & 0.354 [0.285, 0.417]
    & 0.335 [0.280, 0.393] \\
Gemini 2.5 Flash
    & 0.326 [0.264, 0.382]
    & 0.291 [0.240, 0.345] \\
Qwen3-235B
    & 0.373 [0.311, 0.428]
    & 0.316 [0.262, 0.371] \\
\bottomrule
\end{tabular}
\end{minipage}
\hfill
\begin{minipage}[t]{0.37\textwidth}
\centering
\footnotesize
\textbf{(b) Task 2: Next-post prediction}

\vspace{3pt}

\begin{tabular}{@{}lc@{}}
\toprule
\textbf{Method / Model}
& \textbf{Accuracy $\uparrow$} \\
\midrule

\multicolumn{2}{@{}l}{\textit{Baselines}} \\
Uniform random & 0.250 \\
Most frequent answer position & 0.284 \\
Longest candidate & 0.247 \\
Shortest candidate & 0.321 \\
TF--IDF cosine similarity & \textbf{0.556} \\

\midrule
\multicolumn{2}{@{}l}{\textit{Frontier models}} \\
GPT-5.4
    & \textbf{0.877} [0.802, 0.938] \\
Claude Opus 4.6
    & 0.852 [0.765, 0.926] \\
Grok 4.20
    & 0.827 [0.741, 0.901] \\
DeepSeek V3.2
    & \textbf{0.877} [0.802, 0.938] \\
Gemini 2.5 Flash
    & 0.864 [0.790, 0.938] \\
Qwen3-235B
    & 0.827 [0.741, 0.901] \\
\bottomrule
\end{tabular}
\end{minipage}

\caption{Baselines and frontier-model performance on \textsc{CrowdMath} evaluation tasks. Task 1 evaluates post-role classification using macro-F1 and accuracy over four classes: progress, result, erroneous, and find\_error. Task 2 evaluates 4-way multiple-choice next-post prediction using accuracy. Values in brackets are 95\% bootstrap confidence intervals from 10,000 instance-level resamples.}
\label{tab:main_results}
\end{table*}

\begin{figure*}[t]
\centering
\includegraphics[width=\linewidth]{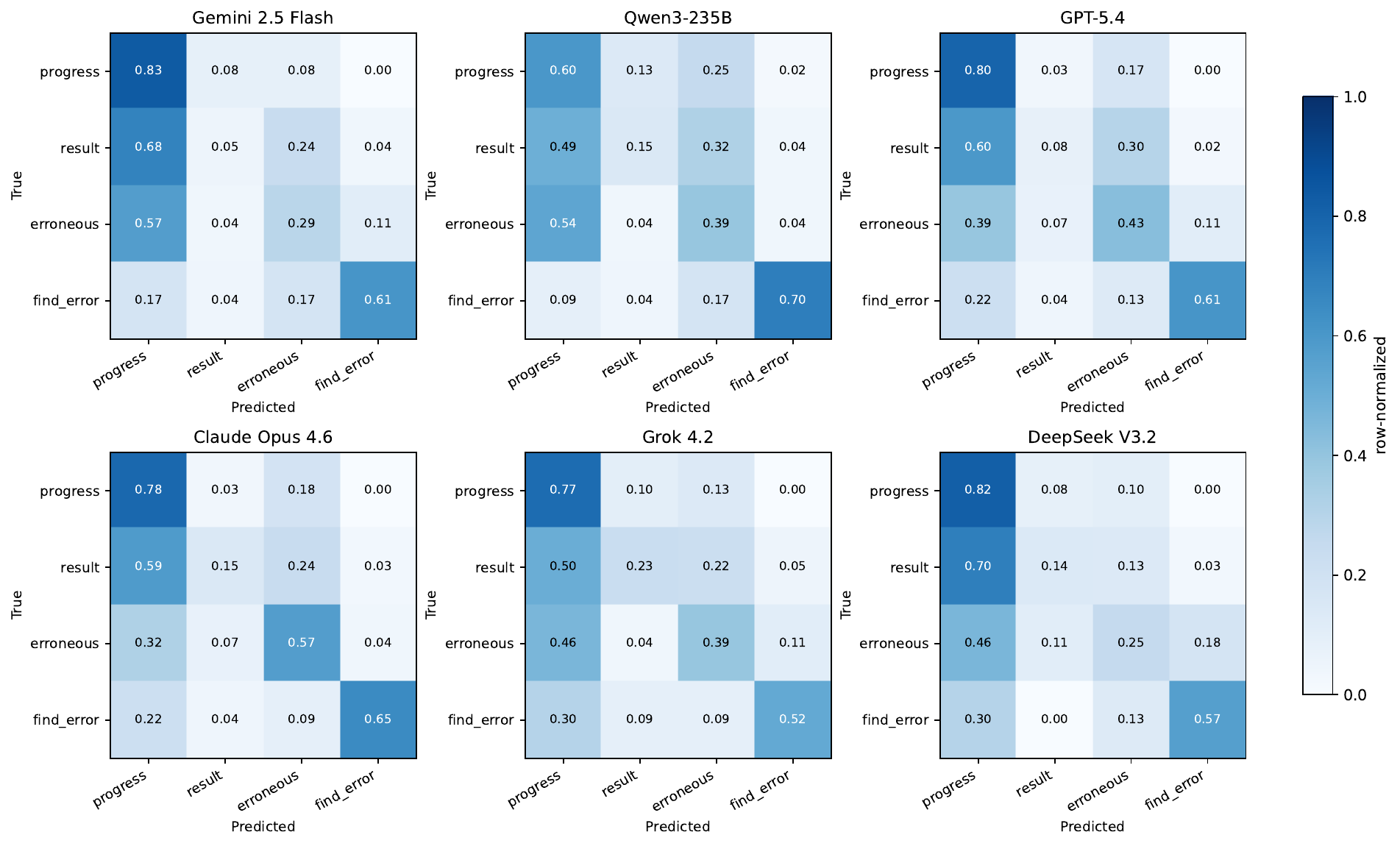}
\caption{Row-normalized confusion matrices for Task 1 post-role classification across six frontier models. Each cell shows the fraction of posts with a given true label (rows) predicted as each class
(columns) normalized along the row.
Values along the diagonal are per-class recall.}
\label{fig:task1-confusion}
\end{figure*}

\subsection{Frontier models struggle to identify proof completions.} To track how well frontier models predict the right post label in our dataset, we construct 275 instances from the dataset. Each \textsc{erroneous} and \textsc{find\_error}
post becomes one instance; \textsc{result} instances are the \textsc{Result} post of each proof
chain; \textsc{progress} instances are sampled one per chain from posts labeled
\textsc{Start}, \textsc{Progress}, or \textsc{NewProgress}. The label counts are 164, 60, 28, and 23 for
\textsc{result} / \textsc{progress} / \textsc{erroneous} / \textsc{find\_error} respectively. The prompts used for this task are in Appendix \ref{app:prompts} and contains one calibration example per class.

The first row of Table \ref{tab:main_results} shows the macro-F1 score across different frontier models. F1 scores are between 0.33 - 0.42, showing that frontier models struggle to identify the different steps in a proof. We report 95\% bootstrap confidence intervals from 10,000 instance-level resamples.  As baselines, we consider predicting the majority class (\textsc{Result}) and a position-only logistic regression classifier. We train the logistic regression clasisfier to predict the four labels from the number of preceding posts in the context, using 5-fold stratified cross-validation grouped by chain. The position-only baseline obtains macro-F1 scores of 0.186, showing that post position alone is insufficient to predict contribution type. Figure \ref{fig:task1-confusion} shows the breakdown of model predictions across all 4 classes. Most models struggle to identify proof completions: rather than labeling these posts as result, they label them as progress. Precision on result is high, showing that when a model labels a post as result it almost always does complete the proof, but recall is low.

\subsection{Frontier models perform well on next-post prediction.}
To measure how well frontier models identify the next post in an ongoing mathematical discussion, we construct 81 multiple-choice instances from the dataset. Each proof chain with at least two posts is considered. All posts before the target post are provided as thread prefix to the model and the model needs to predict the next post out of the 4 options. The remaining three distractor posts are sampled uniformly from other chains in
the same year; chains with fewer than three same-year candidates are dropped. For the target post we prefer sampling posts labeled Start or Progress; if no such posts exist in the chain,
we sample a post at random. Quote blocks are stripped from all texts because they would otherwise make the task trivial. Across 81
instances the target label distribution is 24 Progress, 13 Question, 8 FindError, 5 Answer, 3 Start, 2 NewProgress, 1 Erroneous, and 27 Proof+Result (of which 2 also carry FindError). We evaluate the same six frontier models at temperature 0 and other default settings in OpenRouter for these models with a 1-shot prompt containing one calibration example. The prompts used for this task are in Appendix \ref{app:prompts}.

From table \ref{tab:main_results}, we see that all six models perform well above the 25\% random baseline, with accuracy ranging from 82.7\% to 87.7\%, suggesting that frontier models can perform reasonably well at predicting the next post in a mathematical discourse. We report 95\% bootstrap confidence intervals from 10,000 instance-level resamples. We additionally compare against several simple baselines. The answer-position counts are A:18, B:17, C:23, and D:23, and the baselines select the most frequent answer position, the longest or shortest candidate, or the candidate with the highest TF-IDF cosine similarity to the discussion prefix. For the TF-IDF baseline, we fit a TF-IDF vectorizer using scikit-learn with default settings on the corpus containing the prefixes and four candidate posts for all 81 instances, and select the candidate with the highest cosine similarity to the prefix. The strongest baseline, TF-IDF cosine similarity, achieves 55.6\% accuracy, substantially below the performance of the frontier models.

\subsection{Qualitative Error Analysis}
\textbf{Identifying proof completions:} To understand why RESULT posts over the six models are misclassified, we analyze the models' self-reported justifications (the reasoning field elicited by the prompt) using keyword patterns (we spot-check few of them manually) The patterns are not mutually exclusive.

\begin{enumerate}
    \item Among the 582 RESULT to PROGRESS justifications across all frontier models, in 30\% of these errors, the model judges the post against the wrong target: it says the post "does not fully resolve the original problem." and does not consider that the result label is meant to be a self contained finding. For 17\% of the posts, the model penalizes the informal writing style of the posts, calling the argument "not rigorous," "unproven," or "plausible but," even when the underlying mathematics is complete and correct.

    \item Models label RESULT posts as ERRONEOUS in 28\% of the misclassifications. In 89\% of these ERRONEOUS labels, the model points to a specific mistake it believes it found. In over half of them (55\%), the post comes from a chain whose results were later published in a peer-reviewed paper.

    \item Finally, we check whether a model could identify RESULT posts simply by looking for phrases like "this completes the proof" or "QED." In 90\% of the RESULT posts in the benchmark contain no such phrase, and of the 13 RESULT posts most models classify correctly, only 2 contain such a phrase. Thus, there is no wording shortcut and the model must reason about the actual mathematics in the post.

\end{enumerate}
\textbf{Next Post prediction} We pool the errors across all models and analyze the models' self-reported justifications. The prompt elicits an assessment of every candidate, so for each error we can read why the model rejected the true next post. We use keyword patterns over these assessments, spot-checked a few manually and the categories overlap.
\begin{enumerate}
    \item  In 38\% of the errors, the models reject the true next post because it changes direction by opening a new subproblem, asking a forward-looking question, or reporting a relevant external paper.
    \item In 17\% the models reject the true next post as it is too brief or asking for a clarification.
    \item In 9\% the models reject the true next post as mathematically wrong. These are target posts that correct or disagree with an earlier post.
\end{enumerate}




\section{Conclusion}

We presented \textsc{CrowdMath}, a benchmark for evaluating language models on collaborative open-problem solving. Existing mathematical reasoning benchmarks largely evaluate whether a model can produce a correct final answer or complete
proof. In contrast, \textsc{CrowdMath} evaluates whether models can understand the process by which mathematical results emerge: through partial progress, questions, errors, refinements, and proof completion across multi-post
discussions. Our results reveal a gap between solving well-specified mathematical problems
and understanding collaborative mathematical progress. While frontier models
can often predict the local flow of a discussion, they struggle to judge the
role and significance of individual posts, frequently confusing completed proofs with partial
progress. These findings suggest that current models have
not yet fully learned the discourse-level reasoning needed to track how
mathematical ideas develop over time. By shifting evaluation from final-answer correctness to progress-aware
reasoning, \textsc{CrowdMath} opens a new direction for benchmarking
mathematical intelligence. We hope it supports future work on models that can
participate more effectively in open-ended mathematical inquiry.

\section*{Ethical considerations}

\textsc{CrowdMath} is derived from discussion threads from the CrowdMath program,
an initiative created by MIT PRIMES and AoPS. The study was reviewed by the Institutional Review Board(IRB) and determined to be exempt under Category 4 for secondary use research.
The dataset is intended for
studying collaborative mathematical problem-solving and process-level reasoning
in mathematical discourse. Potential risks include misuse of the dataset outside its intended research context. As described in Section 4.1, the data was anonymized before analysis and release:
usernames were replaced with hash strings, exact timestamps were replaced by
chronological post order, and we performed an additional scan of post text to
remove remaining identifying references. Because the original discussions are publicly accessible, anonymization does not fully eliminate re-identification risk. Users of the released dataset should not attempt to link anonymized posts back to original forum accounts or otherwise
infer participant identities. 
The dataset and accompanying code are publicly available under the MIT License.
The original forum participants were not recruited or compensated for this
study as the dataset is derived from pre-existing public CrowdMath discussions.
Expert annotators were recruited based on mathematical background and compensated
for annotation work under standard project/institutional arrangements.

\section*{Limitations}

The dataset has several limitations that bear on its use. First, it is result-centric: chains are built around completed results, so exploratory reasoning that does not contribute to a finalized result (for example, false starts, abandoned ideas, or dead-end computations) is not represented. This makes the dataset tractable but yields a filtered view of the problem-solving process, and the boundary between partial arguments and finished proofs reflects annotator judgment. Second, labels are assigned at the post level. A post that contains several logically distinct components receives all applicable labels, but the annotation does not record which spans of the text support each label, limiting fidelity for fine-grained reasoning tasks. Third, posts can contribute to more than one completed result and therefore appear in multiple chains. This reflects genuine mathematical structure, for example when a lemma supports several results, but means that naive train/test splits at the post or chain level can leak content across partitions; splitting at the level of results or connected components of the post-contribution graph avoids this. Annotations were produced by different annotators across years, with a single annotator per year. For the 2024 subset, we additionally measure inter-annotator agreement using independent double annotations and report them in Appendix \ref{app:annotation}. However, for the remaining years, annotator-specific conventions are confounded with temporal variation, and inter-annotator agreement cannot be used to quantify uncertainty across the full dataset. The label space is coarse, distinguishing roles such as \textsc{Proof}, \textsc{Erroneous}, and \textsc{Question} but not finer structure such as error types or degrees of rigor. Finally, CrowdMath is an online collaborative program aimed at high school and undergraduate participants, so the discourse, error patterns, and levels of rigor may not generalize to other mathematical settings such as professional research or formal proof systems. 

\section*{Acknowledgments}
This work was funded in part by an Amazon AGI research award to Anna Rumshisky. We thank Harold Polo for his contribution to the annotation.

\bibliography{custom}

\clearpage
\appendix

\section{Algorithm}
\paragraph{Progress chain construction}
The raw annotations are assigned at the post level, but many labels are
result-centric: they indicate that a post proposes, advances, proves, or
critiques a particular mathematical result. To construct the dataset used
in our experiments, we convert these flat post annotations into
result-centric progress chains. Each chain corresponds to one completed
result and contains the open-problem statement, any associated setup or
resource text, and the ordered sequence of posts that contribute to that
result.

This conversion is necessary because a single forum thread may contain
multiple results, and a single post may contribute to more than one
result. We therefore first identify all result identifiers referenced by
the annotations, then build one chain per result by collecting all posts
whose labels refer to that result. Posts are ordered by their original
post number, preserving the temporal structure of the discussion. When a
post contributes to a result completed in a different thread, we resolve
the reference using post numbers that are recorded globally.

Algorithm~\ref{alg:progress-chain-construction} gives the simplified
procedure used to construct these progress chains from the annotated
posts.

\subsection{Model Details}
\label{app:model_details}

We evaluate six frontier models through OpenRouter: GPT-5.4,
Claude Opus 4.6, Grok 4.20, DeepSeek V3.2, Gemini 2.5 Flash,
and Qwen3-235B-A22B. For reproducibility, Table~\ref{tab:model_details}
reports the exact OpenRouter slug used for each model, along with the
provider and the closest available official documentation, model card,
system card, or technical report. Several of these API models do not have
archival technical papers; in those cases, we cite official provider
documentation or model cards. All model evaluations are run at temperature
0 with the remaining OpenRouter settings left at their defaults.

\begin{table*}[t]
\centering
\small
\begin{tabularx}{\textwidth}{lllX}
\toprule
Model & Provider & OpenRouter slug & Official documentation / model card \\
\midrule
GPT-5.4
& OpenAI
& \texttt{openai/gpt-5.4}
& OpenAI model documentation: \url{https://developers.openai.com/api/docs/models/gpt-5.4} \\

Claude Opus 4.6
& Anthropic
& \texttt{anthropic/claude-opus-4.6}
& Anthropic system card: \url{https://www-cdn.anthropic.com/14e4fb01875d2a69f646fa5e574dea2b1c0ff7b5.pdf} \\

Grok 4.20
& xAI
& \texttt{x-ai/grok-4.20}
& xAI model documentation: \url{https://docs.x.ai/developers/models/grok-4.20} \\

DeepSeek V3.2
& DeepSeek
& \texttt{deepseek/deepseek-v3.2}
& DeepSeek technical report: \url{https://huggingface.co/deepseek-ai/DeepSeek-V3.2/resolve/main/assets/paper.pdf} \\

Gemini 2.5 Flash
& Google
& \texttt{google/gemini-2.5-flash}
& Google DeepMind model card: \url{https://storage.googleapis.com/deepmind-media/Model-Cards/Gemini-2-5-Flash-Model-Card.pdf} \\

Qwen3-235B-A22B-Instruct-2507
& Qwen
& \texttt{qwen/qwen3-235b-a22b-2507}
& Hugging Face model card: \url{https://huggingface.co/Qwen/Qwen3-235B-A22B-Instruct-2507} \\
\bottomrule
\end{tabularx}
\caption{Models used in our frontier-model evaluations. We report the exact OpenRouter slug used for inference and link to the closest official documentation, model card, system card, or technical report available for each model.}
\label{tab:model_details}
\end{table*}

\label{app:algo}
\begin{algorithm*}[t]
\caption{Progress Chain Construction (Simplified)}
\label{alg:progress-chain-construction}
\begin{algorithmic}[1]

\State \textbf{Input:} $\mathcal{S}$: the annotated posts from the input file,
       each post carrying its full label set
\State \textbf{Output:} for each result $y$ in $\mathit{ResultSet}$:
       a progress chain $\mathrm{Chain}(y)$ with fields
       \texttt{problem\_text}, \texttt{problem\_resources}, and \texttt{posts}

\StepHeader{Step 1 — Collect all results referenced in the input}

\State $\mathit{ResultSet} \gets \emptyset$
\For{each post $s \in \mathcal{S}$}
  \For{each label $\ell$ on $s$ that references a result $y$}
    \State add $y$ to $\mathit{ResultSet}$
  \EndFor
\EndFor

\StepHeader{Step 2 — Build the chain for each result $y \in \mathit{ResultSet}$}

\For{each $y \in \mathit{ResultSet}$}
  \State $\mathit{Posts}(y) \gets \{s \in \mathcal{S} \mid \text{some label on } s \text{ references } y\}$
  \State $\mathit{ProbPost}(y) \gets$ post in $\mathit{Posts}(y)$ carrying $\mathrm{Problem}(y)$, if any
  \State $t\text{-}n \gets$ open-problem ref carried by any post in $\mathit{Posts}(y)$, if any

  \If{$t\text{-}n$ found}
    \State $\mathit{Setup} \gets$ \textbf{Setup} section(s) from $\mathtt{json\_object\_data}[t\text{-}n]$, if any
    \State $\mathit{ProbSecs} \gets$ \textbf{Problem} section(s) from $\mathtt{json\_object\_data}[t\text{-}n]$, if any
    \State $\mathit{Resources} \gets$ \textbf{Relevant Resources} sections
           from $\mathtt{json\_object\_data}[t\text{-}n]$, if any
    \State $\texttt{problem\_text}(y) \gets \mathit{Setup} \mathbin{\|} \mathit{ProbSecs} \mathbin{\|} \mathit{ProbPost}(y)\mathtt{.text}$
    \State $\texttt{problem\_resources}(y) \gets \mathit{Resources}$ if non-empty, else \textbf{null}
  \Else
    \State $\texttt{problem\_text}(y) \gets \mathit{ProbPost}(y)\mathtt{.text}$ if $\mathit{ProbPost}(y)$ exists, else \textbf{null}
    \State $\texttt{problem\_resources}(y) \gets \textbf{null}$
  \EndIf

  \State $\mathit{Rest}(y) \gets \mathit{Posts}(y) - \{\mathit{ProbPost}(y)\}$
         sorted in ascending order of $\mathit{post\_number}$
  \State $\mathrm{Chain}(y) \gets \{\texttt{problem\_text}(y),\; \texttt{problem\_resources}(y),\; \mathit{Rest}(y)\}$
\EndFor

\end{algorithmic}
\end{algorithm*}

\section{Research Relevance}

The dataset reflects CrowdMath’s focus on research-level problems approachable to high-school and undergraduate students. `Result` denotes a previously unknown finding established in the course of discussion. 88/164 chains are linked to 12 papers, including in the Electronic Journal of Combinatorics and Discrete Applied Mathematics. Of the 88 chains linked to 12 published papers, 22 concern extremal functions of forbidden sequences (3 papers), 21 pursuit–evasion games on graphs (1), 16 metric dimension of graphs (1), 10 graph coloring (1), 7 zero forcing on graphs (1), 6 factorization invariants (1), 3 neural codes on graphs (1), 2 arithmetic of power monoids (2), and 1 powerful numbers (1).

\section{Annotation Quality}
\label{app:annotation}

Two independent experts double-annotated all 298 posts in the 2024 subset.
Both have combinatorics expertise and close familiarity with the threads.
We compare exact per-post canonical label sets for the six core types (Progress, Proof, Question, Answer, FindError, and Erroneous), including arguments and referenced IDs. Post-level agreement is appropriate because chains are generated deterministically from these core labels. Across all posts, 294 agreed and 4 disagreed
($P_o = 98.66\%$, $P_e = 0.8800$, Cohen's $\kappa = 0.8882$).
Among the 20 posts receiving a core label from either annotator,
16 agreed ($P_o = 80.00\%$, $P_e = 0.0650$, $\kappa = 0.7861$).
The same four posts account for the disagreements in both analyses. 
One is a Progress-versus-Answer disagreement, although both link it to the same result.
The other three occur in the same thread; two may reflect a one-post numbering offset. No disagreement involved Proof, no disagreement was Proof versus Progress. These results show strong agreement for 2024

\section{Evaluation Prompts}
\label{app:prompts}

We report the prompts used for post-role classification and 
next-post prediction. Template variables shown as \texttt{\{variable\_name\}} are
substituted with instance-specific content at evaluation time. Both tasks are
evaluated at temperature~0 with default settings via OpenRouter. Post-role classification uses a 4-shot
prompt (one calibration example per class); Next-post prediction uses a 1-shot prompt.

\subsection{Post-Role Classification}
  \label{app:prompt-task2}

  The model is given all preceding posts in a proof chain as context and must classify
  a single target post into one of four functional roles: \textsc{Progress},
  \textsc{Result}, \textsc{Erroneous}, or \textsc{FindError}. The prompt asks the model
  to reason through three explicit steps before producing a structured JSON output.

  \subsubsection*{Prompt template}

  \begin{lstlisting}[style=prompt]
  Proof Step Classification

  You are an expert in mathematical reasoning and discourse analysis. You will be given
  a sequence of posts from a collaborative mathematics research thread, followed by a
  single post to classify.

  <category_definitions>
  Classify the post into exactly one of these four categories:

  PROGRESS - The post makes a meaningful mathematical contribution toward the result
  but does not complete the proof. This includes: introducing a useful lemma, or
  observation, proposing a viable strategy; narrowing the problem; or providing a
  partial argument that advances the thread.

  RESULT - The post provides a complete proof of the result, or supplies the final
  missing step that, together with prior posts, constitutes a full proof.

  ERRONEOUS - The post itself contains a fatal mathematical error or incorrect claim.

  FIND_ERROR - The post identifies a fatal mathematical error in a previous post.
  The current post is correct -- its contribution is to catch a fatal flaw in earlier
  reasoning.
  </category_definitions>

  <instructions>
  Evaluate the post by working through these steps in order:

  Step 1 -- Summarise the post
  In one sentence, state what mathematical claim or action the post performs.

  Step 2 -- Assess against the prior context
  Using only the preceding posts as context, determine:
    - Does this post introduce new correct mathematical content that advances the
  proof?
    - Does it complete the proof given what came before?
    - Does it contain a fatal error of its own?
    - Does it identify a fatal error in a prior post?

  Step 3 -- Assign category
  Apply the definitions above to assign exactly one of:
  PROGRESS, RESULT, ERRONEOUS, FIND_ERROR.

  If the post both finds a fatal error and makes new progress, assign FIND_ERROR --
  surfacing the fatal error is the primary contribution since it invalidates prior
  work.
  If the post contains a fatal error but also partially advances the thread, assign
  ERRONEOUS.
  </instructions>

  <output_format>
  Provide your classification inside <classification> and </classification> tags using
  this JSON structure exactly with no text outside the tags:
  {
    "summary": "<one sentence describing what the post does>",
    "reasoning": "<2-3 sentences explaining why this category applies,
                   referencing the post content and prior context>",
    "label": "progress" | "result" | "erroneous" | "find_error"
  }
  </output_format>

  <calibration_examples>
  [Four examples, one per class -- see below.]
  </calibration_examples>

  Prior context:
  {prior_context}

  Post to classify:
  {question}

  <judgement>
  </judgement>
  \end{lstlisting}

  \subsubsection*{Calibration examples}

  The prompt includes one example per class, drawn from the 2016 pattern-avoidance
  thread. We show all four below.

  \begin{lstlisting}[style=prompt]
  Example 1 -- RESULT:

  <post>
  If $n$ is even and at least one of $j$ and $k$ is odd, player 2 wins. Say $k$ is
  odd, then after player 1 flips $(i, 2d+1)$, player 2 flips $(i, 2d+2)$, or vice
  versa. So after each move of player 2, column $2d+1$ and $2d+2$ are identical for
  all $d$, and the matrix cannot start to contain $P$ which has an odd number of
  identical columns after player 2's move.
  So, to generalize: if $n$ is even and $P$ is composed of an odd number of identical
  rows or columns, $W(n, P)=2$.
  </post>

  <judgement>
  {
    "summary": "The post proves that Player 2 wins when n is even and P has an odd
                 number of identical rows or columns, by constructing an explicit
                 mimic strategy.",
    "reasoning": "Post #13 asked whether a mimic strategy could prove W(n,P)=2 for
                  certain cases. This post answers that question completely: it
                  constructs the strategy (mirror columns 2d+1 and 2d+2), proves it
                  keeps the matrix P-free, and states the general result. The proof
                  is self-contained and contains no gaps.",
    "label": "result"
  }
  </judgement>

  Example 2 -- PROGRESS:

  <post>
  It depends on what exactly "this strategy" is, in particular the way of mimic
  employed by player 2. If player 2 always flips $(i, 2d+1)$ after player 1 flips
  $(i, 2d+2)$ and vice versa, then a pattern $P$ with even number of columns, along
  with column 1 identical to column 2, column 3 to column 4, etc., guarantees victory
  for player 1.
  As for all the matrices $P$ for which player 2 can guarantee victory with this
  strategy when $n$ is even, I haven't got better idea than those with odd number of
  identical rows or columns.

  I also wonder for an odd $n$, suppose the center of $n$-by-$n$ matrix has coordinate
  $(0, 0)$. What patterns lead to player 1's victory if they first flip $(0, 0)$ and
  then flips $(-i, -j)$ after player 2 flips $(i, j)$?
  </post>

  <judgement>
  {
    "summary": "The post gives a partial answer to the characterization question by
                 identifying one class of matrices where Player 1 wins under the mimic
                 strategy, but does not complete the characterization.",
    "reasoning": "Post #16 asked for a full characterization of matrices P where
                  Player 2 wins. This post advances that question by exhibiting a
                  concrete counter-class (even columns with paired identical columns
                  gives Player 1 a win), but explicitly admits it has no general
                  characterization beyond the odd-identical-columns case already known.
                  The proof is not complete -- the thread remains open.",
    "label": "progress"
  }
  </judgement>

  Example 3 -- ERRONEOUS:

  <post>
  A simple thought. All player indices below are $0$-based.
  For an even $k$, we should have loser $L_2(n,P) \equiv L_k(n,P) \pmod{2}$, where
  the subscript means the number of players. This is because if $L_2(n,P)=1$, then
  players $0, 2, 4, \ldots, k-2$ can collectively deploy player $0$'s winning strategy
  as if there are only 2 players to make sure not to lose.
  To generalize, $L_d(n,P) \equiv L_{md}(n,P) \pmod{d}$.
  </post>

  <judgement>
  {
    "summary": "The post claims that players can collectively deploy a two-player
                 winning strategy in a k-player game, and generalizes this to
                 L_d(n,P) = L_{md}(n,P) mod d.",
    "reasoning": "The argument assumes players 0, 2, 4, ..., k-2 will cooperate to
                  execute player 0's two-player strategy, but gives no justification
                  for why the other players would do so -- each player acts
                  independently and may deviate. The collective strategy argument is
                  unverified, making the generalization unsupported.",
    "label": "erroneous"
  }
  </judgement>

  Example 4 -- FIND_ERROR:

  <post>
  [Quoting post #24's argument.]
  Is there an issue if any players among $0, 2, 4, \ldots, k-2$ do not deploy
  player 0's two-player winning strategy?

  Also, what happens for $md$ players when there are multiple $d$-player winning
  strategies?
  </post>

  <judgement>
  {
    "summary": "The post identifies two gaps in post #24's collective-strategy
                 argument: non-cooperative players and ambiguity when multiple winning
                 strategies exist.",
    "reasoning": "The post correctly pinpoints that post #24's proof requires all
                  grouped players to follow the same strategy, which is not guaranteed
                  since players act independently. It also raises the valid concern
                  that the argument breaks down when the d-player winning strategy is
                  not unique. This post contains no errors of its own -- it is a
                  correct critique of prior reasoning.",
    "label": "find_error"
  }
  </judgement>
  \end{lstlisting}

  \subsection{Next-Post Prediction}
  \label{app:prompt-task3}

  The model is given a thread prefix and four candidate posts (A--D). Exactly one is
  the true next post; the other three are distractors drawn from different chains in
  the same year. The model reasons in three steps before selecting a candidate. A
  single
  calibration example is included in the prompt.

  \begin{lstlisting}[style=prompt]
  You are an expert in mathematical reasoning and discourse analysis. You will be given
  a research problem, a thread prefix showing the discussion so far, and four candidate
  posts (A, B, C, D). Exactly one is the actual next contribution to the proof. The
  other three are real posts drawn from different threads and are distractors.

  Your task is to identify which candidate is the most likely next post in this thread.

  <instructions>
  Work through these steps in order before selecting your answer.

  Step 1 -- Summarise the proof state
  In 2-3 sentences, describe what has been established in the thread so far.

  Step 2 -- Evaluate each candidate
  For each of A, B, C, D, write one sentence assessing whether it fits as the next
  post.
  Check:
  - Does it address an open question or gap identified in the thread prefix?
  - Does it use notation and concepts consistent with the thread?
  - Is it at the right level -- neither repeating what is already done nor jumping past
    the current stage of the proof?

  Step 3 -- Select the best candidate
  Choose the single candidate that best fits as the next post. If multiple candidates
  seem plausible, select the one that best fits the evaluation criteria.
  </instructions>

  <success_criteria>
  The correct next post:
  1. Follows directly from what the thread prefix has established or asked.
  2. Uses notation already introduced in the thread.
  3. Makes progress at a level consistent with the thread's current stage.
  4. Does not introduce a topic unrelated to the current proof direction.
  </success_criteria>

  <output_format>
  Provide your selection inside <selection> and </selection> tags using this JSON
  structure exactly -- no text outside the tags:
  {
    "proof_state": "<2-3 sentences summarising what is established>",
    "candidate_assessments": {
      "A": "<one sentence>",
      "B": "<one sentence>",
      "C": "<one sentence>",
      "D": "<one sentence>"
    },
    "answer": "A" | "B" | "C" | "D"
  }
  </output_format>

  <calibration_example>
  <thread_prefix>
  For what patterns $P$ is it true that any algorithm to determine whether $A$ contains
  $P$ must have $\Omega(h_A w_A)$ running time?

  Post #31: The direction of this question sounds interesting, but I feel it may need
  some tweak. For $h_A \gg h_P$ and $w_A \gg w_P$, it seems a scan of most of the
  $w_A h_A$ elements are required. We can always have an $A$ with most of the elements
  $0$, and yet still contains $P$. Scanning the $\approx h_A w_A$ zeros seems
  inevitable when we are unlucky.

  Post #33: Is it true that any algorithm to determine whether $A$ contains $P$ must
  check almost every entry of $A$ if at least one entry of $P$ is $1$?
  What if $P$ has all zeroes?
  </thread_prefix>

  <candidates>
  A. Lemma: If $n = 2$ and $k = 2$, then $W(n, P)$ will always be Player 2.
     Proof: Let $A$ be the $2 \times 2$ all-zeroes matrix and $P$ be the $2 \times 1$
     all-ones matrix. Player 1 must change one zero to a one. In either scenario for
     Player 2's response, Player 1 is forced to create the forbidden pattern $P$ on
     the next move.

  B. I think it's positive: the number of ones of a minimally non-linear binary matrix
     with $k$ rows is $O(k)$. We can ignore rows with a single one, because they
     contribute $O(k)$ ones. We claim that every column has no more than $2$ ones,
     followed by the fact that there are at most $4k-2$ columns.

  C. If $A$ is $n \times n$, then any algorithm has to check at least $n^2 - Ex(n,P)$
     entries to declare $A$ avoids $P$, because if it skips $Ex(n,P)+1$ elements then
     in the worst case those skipped elements are all $1$ and contain $P$. So if
     $Ex(n,P) = O(n)$ then the proportion of elements that must be checked before
     declaring $A$ avoids $P$ is close to $1$.

  D. I think this family also solves crowdmath open problem 2. The edge metric
  dimension
     of the $k^{th}$ graph in this family is $3^k(1-o(1))$ and the metric dimension is
     $k$. To see why, it suffices to prove: if $G$ has order $n$ and some vertex of
     degree $n-1-x$ within distance $2$ of all vertices, then
     $edim(G) \geq n-1-x-2^x$.
  </candidates>

  <selection>
  {
    "proof_state": "Post #31 observes that scanning most of an n x n matrix seems
      unavoidable when checking for P-avoidance. Post #33 sharpens this into two
      specific questions: must every algorithm check nearly all entries when P has at
      least one 1, and what happens when P is all-zero? The all-zero sub-case and the
      general lower bound remain open.",
    "candidate_assessments": {
      "A": "Proves W(2,2)=Player 2 in a pattern avoidance game -- completely unrelated
            to algorithm complexity or the Ex(n,P) framework of this thread.",
      "B": "Bounds the number of ones in a minimally non-linear matrix -- a structural
            combinatorics result unrelated to the algorithmic question about checking
            time.",
      "C": "Directly answers Post #33's first question by proving a lower bound of
            n^2 - Ex(n,P) checks using a worst-case argument, using Ex(n,P) notation
            already established in the thread.",
      "D": "Claims a result about edge metric dimension and metric dimension of a graph
            family -- from a different thread on a different topic entirely."
    },
    "answer": "C"
  }
  </selection>
  </calibration_example>

  Identify the most likely next post in this thread.

  <thread_prefix>
  {thread_prefix}
  </thread_prefix>

  <candidates>
  A. {candidate_a}

  B. {candidate_b}

  C. {candidate_c}

  D. {candidate_d}
  </candidates>

  <selection>
  </selection>
  \end{lstlisting}

  \newpage
\end{document}